\documentclass[12pt,reqno]{amsart}

\usepackage[T1]{fontenc}
\usepackage{enumitem}
\usepackage{hyperref}
\hypersetup{
  colorlinks   = true,
  citecolor    = blue,
  linkcolor    = blue
}
\usepackage{amsmath,amsfonts,amsthm,amssymb,color,tikz, comment,graphicx,epsfig}
\usepackage{mathrsfs}

\usepackage{pdfsync}
\usepackage[font={scriptsize}]{caption}

\usepackage{romannum}
\AtBeginDocument{\pagenumbering{arabic}}

\usepackage[left=1in, right=1in, top=1.1in,bottom=1.1in]{geometry}
\usepackage[backend=bibtex, style=alphabetic, sorting=nyt]{biblatex}
\newcommand{\dd}{\mathrm{d}}

\newcommand{\ots}{[0,\sigma]}
\newcommand{\ott}{[0,\tau]}

\newcommand{\1}{{\bf 1}}
\newcommand{\2}{{\bf 2}}

\newcommand{\R}{\mathbb R}
\newcommand{\N}{\mathbb N}

\newcommand{\bx}{{\mathbf{x}}}

\newcommand{\cs}{\mathcal S}

\newtheorem{theorem}{Theorem}[section]

\theoremstyle{remark}
\newtheorem{remark}[theorem]{Remark}

\theoremstyle{remark}

\newcommand{\bean}{\begin{eqnarray*}}
\newcommand{\eean}{\end{eqnarray*}}
\newcommand{\ben}{\begin{enumerate}}
\newcommand{\een}{\end{enumerate}}
\newcommand{\beq}{\begin{equation}}
\newcommand{\eeq}{\end{equation}}

\usepackage{pifont}

\begin{document}

\title[2-d signature]{2-d signature of images and texture classification}


\author[S. Zhang]{Sheng Zhang}
\author[G. Lin]{Guang Lin}
\author[S. Tindel]{Samy Tindel}

\address{Sheng Zhang: Department of Mathematics,
Purdue University,
150 N. University Street,
W. Lafayette, IN 47907,
USA.}
\email{zhan2694@purdue.edu}

\address{Guang Lin: Department of Mathematics,
Purdue University,
150 N. University Street,
W. Lafayette, IN 47907,
USA.}
\email{guanglin@purdue.edu}

\address{Samy Tindel: Department of Mathematics,
Purdue University,
150 N. University Street,
W. Lafayette, IN 47907,
USA.}
\email{stindel@purdue.edu}

\begin{abstract}
We introduce a proper notion of 2-dimensional signature for images. This object is inspired by the so-called rough paths theory, and it captures many essential features of a 2-dimensional object such as an image. It thus serves as a low-dimensional feature for pattern classification. Here we implement a simple procedure for texture classification. In this context, we show that a low dimensional set of features based on signatures produces an excellent accuracy.
\end{abstract}

\maketitle

\section{Introduction}

Signatures of paths are fascinating objects which have been under intense scrutiny for the past 25 years. In this introduction we will first put our own investigation into context, summarizing some of the contributions for signatures of paths indexed by a 1-dimensional (1-d) parameter. This will be the content of Section~\ref{sec:intro-1d-signatures}. Then in Section~\ref{sec:intro-2d-signatures} we introduce the concept of signature for paths indexed by a 2-dimensional (2-d) parameter. This is the natural framework in order to deal with image processing. Eventually we will summarize our main results and draw some conclusions.

\subsection{1-d signatures}\label{sec:intro-1d-signatures}

Before going further, let us mention that the \emph{1-d} in  1-d signatures refers to the fact that the paths $x$ we are considering for now are indexed by a 1-d parameter $t\in\ots$ for a given time horizon $\sigma>0$ (as opposed to the 2-d parameter $(s;t)\in\ots\times\ott$ considered in the next sections). Nevertheless, the path $x$ is generally $\R^{d}$-valued, that is $x_{t}\in\R^{d}$ for all $t\in\ots$.

Signatures of paths are prominent objects in analysis and data science due to several remarkable properties: they show up naturally in fundamental computations, they enjoy suitable algebraic and analytic relations, they characterize paths, and they have been successfully applied in a data analysis context. Let us briefly review those features.

\begin{enumerate}[wide, labelwidth=!, labelindent=0pt, label=(\roman*)]
\setlength\itemsep{.1in}
\item
\emph{Prominence of signatures in system approximations.}
One of the simplest situations in which signatures pop up in a very natural way is through basic change of variables formulae.
Namely consider a regular enough $C^{1}$-path (or curve) $x:\ots\to\R^{d}$  and a smooth function $f:\R^{d}\to\R$.
We denote by $x^{i}$ each component of $x$, so that $x=(x^{1},\ldots,x^{d})$. For notational sake, we also write $\partial_{i}f$ for the partial derivative $\partial f / \partial x_{i}$.
Then the most basic form of a change of variable formula asserts that for $0\le s<\hat{s}\le\sigma$ we have
\begin{equation}\label{eq:change-vb-formula}
f(x_{\hat{s}}) - f(x_{s})
=
\sum_{i_{1}=1}^{d} \int_{s}^{\hat{s}} \partial_{i_{1}} f(x_{r_{1}}) \, dx_{r_{1}}^{i_{1}} .
\end{equation}
In order to get further expansions according to relation~\eqref{eq:change-vb-formula}, let us introduce the first two elements of the signature of $x$. They are defined as iterated integrals of $x$  as follows,
\begin{equation*}
\bx_{s\hat{s}}^{\1,i_{1}} = x_{\hat{s}}^{i_{1}} - x_{s}^{i_{1}} = \int_{s<r_{1}<\hat{s}} \dd x_{r_{1}}^{i_{1}}
\quad\text{and}\quad
 \bx_{s\hat{s}}^{\2,i_{1},i_{2}} = \int_{s<r_{1}<r_{2}<\hat{s}} \dd x_{r_{1}}^{i_{1}} \, \dd x_{r_{2}}^{i_{2}} .
\end{equation*}
With this notation in hand and iterating formula~\eqref{eq:change-vb-formula}, we get the following approximation
\begin{equation}\label{eq:approx-change-variable}
f(x_{\hat{s}}) - f(x_{s})
\simeq
\sum_{i_{1}} \partial_{i_{1}} f(x_{s}) \, \bx_{s\hat{s}}^{\1,i_{1}}
+ \sum_{i_{1},i_{2}} \partial_{i_{1},i_{2}}^{2} f(x_{s}) \, \bx_{s\hat{s}}^{\2,i_{1},i_{2}} .
\end{equation}
As one can see from relation~\eqref{eq:approx-change-variable}, the elements $\bx^{\1},\bx^{\2}$  play the role of monomials in a Taylor type expansion along the path $x$. They can thus be thought of as building blocks for a faithful representation of the path $x$.

Extrapolating on this kind of consideration, the signature of $x$ summarizes all the iterated integrals of $x$ in a single object $[S(x)]_{s\hat{s}}$ which can be written as
\begin{equation}\label{eq:signature}
[S(x)]_{s\hat{s}} = 1
+ \sum_{n=1}^{\infty} \int_{s<r_{1}<r_{2}<\cdots<r_{n}<\hat{s}}
\dd x_{r_{1}} \otimes \dd x_{r_{2}} \otimes\cdots\otimes \dd x_{r_{n}} .
\end{equation}
For a given couple $s,\hat{s}$ with $s<\hat{s}$, the element $[S(x)]_{s\hat{s}}$, called the signature of $x$, lies in the space $T(\R^{d})=
\oplus_{n=0}^{\infty} (   \R^{d} )^{\otimes n}$.
It should be mentioned that $S(x)$ also appears very naturally when computing Taylor type expansions of ordinary differential equations. This fundamental property is the one that made signatures the central object in rough paths analysis. Rough paths can be seen as a new point of view on stochastic differential equations, and had a profound impact on stochastic analysis over the past two decades.
Proper generalizations of rough paths are also at the heart of the celebrated regularity structure theory~\cite{Hai14}.

\item
\emph{Algebraic and analytic properties.}
The fact that $S(x)$ belongs to the free algebra $T(\mathbb{R}^d)$ induces very convenient properties for algebraic manipulations. To name just a few, one can prove that if $x\sqcup y$ denotes the concatenation of two paths $x$ and $y$, then (see e.g.~\cite[Theorem 7.11]{FV10}) we have
\begin{equation}\label{eq:sxy}
  S(x\sqcup y) = S(x)\otimes S(y),
\end{equation}
where the product in the right hand side of~\eqref{eq:sxy} is the polynomial type product on $T(\mathbb{R}^d)$. Invariance by reparametrization also holds. More specifically, if $\phi:\ots\to\ots$ is a non-decreasing surjection and if we set $x^\phi = x\circ \phi$, then for all $0\leq s<\hat{s}\leq \sigma$ the following holds true:
\begin{equation}\label{eq:invariance}
  [S(x)]_{\phi(s)\phi(\hat{s})} = S(x^\phi)_{s\hat{s}}.
\end{equation}

On the analytic side, the essential bound on signatures asserts a factorial decay with respect to the order of the integral. Namely, if $S_n(x)$ denotes the $n$-th order integral in~\eqref{eq:signature}, then we have the following upper bound:
\begin{equation}\label{eq:upper-bound}
  ||S_n(x)||\leq \frac{(C_{\tau,x})^n}{n!},
\end{equation}
where the constant $C_{\tau,x}$ does not depend on $n$. Notice that relation~\eqref{eq:upper-bound} yields crucial bounds on differential equations driven by $x$, which can be then extended to stochastic cases.

\item
\emph{Characterization of paths.}
One of the most fundamental properties of signatures (especially with data analysis in mind) is that they characterize paths. Specifically, for two Lipschitz paths $x,y$ we have
\begin{equation}\label{eq:paths}
  S(x)_{01} = S(y)_{01} \qquad\text{iff}\qquad x\sim y,
\end{equation}
where $x\sim y$ means that $x,y$ only differs by a tree-like path. This result is proved in~\cite{HL10}, while~\cite{Gen17} provides an algorithm allowing to reconstruct a path from its signature. Along the same lines, any continuous map $f: C^1([0,\sigma])\to \mathbb{R}$ can be approximated by a linear functional of the signature. This point of view leads to more quantitative versions of~\eqref{eq:paths}, see~\cite{KLA20}.

\item
\emph{Signatures and data analysis.}
Due to the algebraic, analytic and characterization properties recalled above, signatures have been recently used as efficient features in data analysis for paths. The literature on this topic is now abundant. Among those contributions, let us single out the very successful Chinese character recognition algorithm~\cite{Gra13}. Other significant applications include Finance time series~\cite{LNO14}, topological data analysis~\cite{CNO20} and diagonosis prediction~\cite{MLG19}. It is fair to claim that signatures are now accepted as an efficient set of features for classification or prediction of paths.
\end{enumerate}

\subsection{2-d signatures}\label{sec:intro-2d-signatures}
Although there are now promising steps in the analysis of signatures for fields indexed by $m$-dimensional rectangles, this area of research is still in its infancy. Following the structure of Section~\ref{sec:intro-1d-signatures}, let us recall what has been done in case of fields indexed by $\ots\times\ott$.

\begin{enumerate}[wide, labelwidth=!, labelindent=0pt, label=(\roman*)]
\setlength\itemsep{.1in}
\item
\emph{Signatures for calculus in the plane.}
Let us start from the equivalent of relation~\eqref{eq:change-vb-formula} in the plane. Namely consider a $C^2$-field $x:\ots\times\ott\to\R^d$ and a smooth function $f:\R^d\to\R^d$. In order to write more compact equations below, we will use the following conventions:
\begin{equation}\label{eq:conventions}
  \dd_{12} x_{s;t}^i = \frac{\partial^2 x_{s;t}^i}{\partial s \partial t} \, \dd s \dd t,
  \quad\text{and}\quad
  \dd_{\hat{1}\hat{2}} x_{s;t}^{ij} = \frac{\partial x_{s;t}^i}{\partial s} \frac{\partial x_{s;t}^j}{\partial t} \, \dd s \dd t.
\end{equation}
Also recall that we write $\partial_i f$ for the partial derivatives of $f$. In addition, increments like $f(x_{\hat{s}})-f(x_s)$ in~\eqref{eq:change-vb-formula} should be replaced by rectangular increments. Specifically, for a rectangle $[s,\hat{s}]\times[t,\hat{t}]$ and a field $y$ defined on $\ots\times\ott$ we write
\begin{equation}\label{eq:box}
  \Box_{s\hat{s};t\hat{t}} \, y = y_{\hat{s};\hat{t}} - y_{s;\hat{t}} - y_{\hat{s};t} + y_{s;t}.
\end{equation}
Then the equivalent of~\eqref{eq:change-vb-formula} in rectangles gives a change of variables formula for the rectangular increments of $y=f(x)$. It reads
\begin{multline}\label{eq:change-vb-formula-2d}
  \Box_{s\hat{s};t\hat{t}} f(y)
  = \sum_{i,j=1}^{d}\int_{s<s_1<\hat{s}}\int_{t<t_1<\hat{t}}
  \partial_{ij}^2 f(x_{s_1;t_1}) \, \dd_{\hat{1}\hat{2}} x_{\hat{s}_1;\hat{t}_1}^{ij} \\
  + \sum_{i=1}^{d}\int_{s<s_1<\hat{s}}\int_{t<t_1<\hat{t}}\partial_i f(x_{s_1;t_1})
  \,  \dd_{12} x_{\hat{s}_1;\hat{t}_1}^i.
\end{multline}
With respect to~\eqref{eq:change-vb-formula}, the appearance of the term $\dd_{\hat{1}\hat{2}} x$ in~\eqref{eq:change-vb-formula-2d} is obviously a major difference. As a result when one tries to push forward Taylor type expansions up to order 2 like in~\eqref{eq:approx-change-variable}, the number of terms explodes. A description of those terms is given in~\cite{CG14,CT15} for second order calculations, but we are not aware of extensions up to an arbitrary order. Therefore obtaining an expression for a 2-d signature $S(x)$ similar to~\eqref{eq:signature} is still a challenging open problem. Let us also mention that the stochastic calculus for plane-indexed processes does not fit into the general regularity structure framework~\cite{Hai14}, due to some boundary type singularities.

\item
\emph{Algebraic and analytic properties.}
Since the very definition of $S(x)$ for paths indexed by $(s,t)$ is still open, the literature on its algebra is obviously scarce. The recent preprint \cite{Giu+22} gives an account on the algebraic structure generated by the increments
\begin{equation}\label{eq:algebraic-structure-2d}
  \int_{s<s_1<\dots<s_n<\hat{s}}\int_{t<t_1<\dots<t_n<\hat{t}} \dd_{\hat{1}\hat{2}} x_{s_1;t_1} \otimes \dots \otimes \dd_{\hat{1}\hat{2}} x_{s_n;t_n}.
\end{equation}
Some potential generalization of~\eqref{eq:sxy}-\eqref{eq:invariance}-\eqref{eq:upper-bound} are provided therein. In particular, the invariance property~\eqref{eq:invariance} is restricted to coordinate-wise changes of variables of the form
\begin{equation}
  \phi(s,t) = \phi_1(s) \phi_2(t),
\end{equation}
with two non-decreasing surjections $\phi_1$ and $\phi_2$.

\item
\emph{Characterization of paths.}
Here again, only partial information for 2-d signatures is available. However,~\cite[Theorem 6]{Giu+22} is encouraging. Indeed this theorem claims that signatures generated by~\eqref{eq:algebraic-structure-2d} do characterize paths, albeit in a topological weak sense.

\item
\emph{Signature and data analysis.}
A \textsc{rgb} image can be fairly well represented by a field $x:\ots\times\ott\to\R^3$, where each $x_{s;t} = (x_{s;t}^1, x_{s;t}^2, x_{s;t}^3)$ represents a pixel and every coordinate stands for a fundamental color (say $1=$ red, $2=$ green, $3=$ blue). Nevertheless, to the best of our knowledge, 2-d signatures have not been used as features in image processing or other multiparametric data analysis problems. The need for nonlinear functionals of the field $x$ in texture classification was acknowledged in the influential paper~\cite{SM14}, leading to a wide variety of generalizations. One can also mention the recent contribution~\cite{CGW21}, where regularity structures based features are used for prediction purposes. However, the fact that 2-d signatures are natural objects to consider for image processing is not mentioned in those two references.
\end{enumerate}

\subsection{Outline}
With the above preliminaries in mind, our objective can be summarized as follows: we wish to show empirically that 2-d signatures are natural are efficient low-dimensional features for image processing. We propose to achieve this by considering a simple texture identification problem. We will see that considering a 12-dimensional feature and applying standard classification methods, one can achieve an excellent accuracy. This certainly calls for further developments, which will be highlighted in our concluding remarks.

Our paper is structured as follows: in Section~\ref{sec:description-signatures} we give a mathematical description of the signatures we are using for classification purposes, together with their discretized versions. Section~\ref{sec:numerical-experiment} focuses on the numerical experiment, as well as the outcome in terms of accuracy for our classification task. We finish the paper with some concluding remarks in Section~\ref{sec:conclusion}.

\section{Description of the 2-d signatures}\label{sec:description-signatures}
This section is devoted to introduce the features we advocate for in this paper. In Section~\ref{sec:features-continous} we define those objects in the (continuous) plane, while Section~\ref{sec:discretization} focuses on the corresponding discrete objects.

\subsection{Features in continuous space}
\label{sec:features-continous}
The integrals considered below are based on a notion of simplex in the plane. The first of these objects is simply a rectangle of the form $[s,\hat{s}]\times [t,\hat{t}]$. It is denoted below as
\begin{equation}\label{eq:signature-rectangle-1}
  \cs_{1}(s,\hat{s};t,\hat{t}) = \{(s_1,t_1)\in\R^2, s<s_1<\hat{s}, t<t_1<\hat{t}\}.
\end{equation}
Notation~\eqref{eq:signature-rectangle-1} can then be easily extended in order to describe the simplex used for second order integrals. Specifically, we integrate over domains of the form
\begin{equation}\label{eq:signature-rectangle-2}
  \cs_{2}(s,\hat{s};t,\hat{t}) = \{(s_1,s_2,t_1,t_2)\in\R^4; \, s<s_1<s_2<\hat{s}, t<t_1<t_2<\hat{t}\}.
\end{equation}

With the above notation~\eqref{eq:signature-rectangle-1} in hand, the first order increments used as features can be written as
\begin{equation}\label{eq:increments-1}
  \bx_{s\hat{s};t\hat{t}}^{(1,2);i_1} = \int_{\cs_{1}(s,\hat{s};t,\hat{t})} d_{12} x_{s_1;t_1}^{i_1}
  \quad\text{and}\quad
  \bx_{s\hat{s};t\hat{t}}^{(\hat{1},\hat{2});i_1} = \int_{\cs_{1}(s,\hat{s};t,\hat{t})} d_{\hat{1}\hat{2}} x_{s_1;t_1}^{i_1}.
\end{equation}
Notice that $\bx_{s\hat{s};t\hat{t}}^{(1,2);i_1}$ above is simply the rectangular increment $\Box_{s\hat{s};t\hat{t}}\,x$ introduced in~\eqref{eq:box}. Taking into account the fact that the color index $i_1$ lies in $\{1,2,3\}$, we get 6 features of order 1 defined by~\eqref{eq:increments-1}.

Let us now turn to the definition of second order increments. Recalling notation~\eqref{eq:signature-rectangle-2}, we now have several possibilities combing spatial differentials and color indices. We get
\begin{equation}\label{eq:increments-2}
  \bx_{s\hat{s};t\hat{t}}^{(11,22);i_1,i_2} = \int_{\cs_{2}(s,\hat{s};t,\hat{t})} d_{12} x_{s_1;t_1}^{i_1} d_{12} x_{s_2;t_2}^{i_2}
  \quad\text{and}\quad
  \bx_{s\hat{s};t\hat{t}}^{(\hat{1}\hat{1},\hat{2}\hat{2});i_1,i_2} = \int_{\cs_{2}(s,\hat{s};t,\hat{t})} d_{\hat{1}\hat{2}} x_{s_1;t_1}^{i_1} d_{\hat{1}\hat{2}} x_{s_2;t_2}^{i_2}.
\end{equation}

\begin{remark}
  A wider variety of second order increments is available, when one mixes the $d_{12}$ and $d_{\hat{1}\hat{2}}$ differentials defined by~\eqref{eq:conventions}. One can also decide to integrate in one direction only for some of the integrals. See~\cite[p. 5 and p. 18]{CT15} for an exhaustive first of necessary increments for a rough integration in the plane. From this long list of possible increments, we will also appeal to $\bx^{1\hat{1};2\hat{2}}$ and $\bx^{\hat{1}1;\hat{2}2}$, respectively, defined by
\begin{equation}\label{eq:increments-3}
  \bx_{s\hat{s};t\hat{t}}^{(1\hat{1},2\hat{2});i_1 i_2} = \int_{S_2(s,\hat{s};t,\hat{t})} d_{12} x_{s_1;t_1}^{i_1} d_{\hat{1}\hat{2}} x_{s_2;t_2}^{i_2}
  \quad\text{and}\quad
  \bx_{s\hat{s};t\hat{t}}^{(\hat{1}1,\hat{2}2);i_1 i_2} = \int_{S_2(s,\hat{s};t,\hat{t})} d_{\hat{1}\hat{2}} x_{s_1;t_1}^{i_1} d_{12} x_{s_2;t_2}^{i_2}.
\end{equation}
\end{remark}

\begin{remark}
  In this article we focus on low dimensional features. We will thus only consider the increments in~\eqref{eq:increments-2} for $i_1=i_2$. Specifically, the collection of first and second order increments used below will be
  \begin{equation}\label{eq:increments-all}
    \left\{x^{(1,2);i}, x^{(\hat{1},\hat{2});i}, x^{(11,22);i,i}, x^{(\hat{1}\hat{1},\hat{2}\hat{2});i,i}, x^{(1\hat{1},2\hat{2});i,i},  x^{(\hat{1}1,\hat{2}2);i,i}; i=1,2,3\right\}.
  \end{equation}
  We thus end up with a 18-dimensional feature space.
\end{remark}

\begin{remark}
  As mentioned in the introduction, the collection~\eqref{eq:increments-all} of iterated integrals enjoys some complex algebraic properties. We refer to~\cite[Section 5.1]{CG14} for an account on these relations. See also the aforementioned article~\cite{Giu+22}.
\end{remark}

\begin{remark}
  Analytic properties are  part of the appeal of signatures as features. Since our considerations in the current article are restricted to second order increments, the factorial decay exhibited in relation~\eqref{eq:upper-bound} does not really make sense. However, assuming that $x$ is a $C^2$-signal, it is readily checked that for all $0\le s<\hat{s}\le \sigma$ and $0\le t<\hat{t}\le \tau$ we have
  \begin{equation}
    \begin{aligned}
    \left|x_{s\hat{s};t\hat{t}}^{(1,2),i}\right| + \left|x_{s\hat{s};t\hat{t}}^{(\hat{1},\hat{2}),i}\right| &\le C_x \left|\hat{s}-s\right| \left|\hat{t}-t\right| \\
    \left|x_{s\hat{s};t\hat{t}}^{(11,22),ii}\right| + \left|x_{s\hat{s};t\hat{t}}^{(\hat{1}\hat{1},\hat{2}\hat{2}),ii}\right| + \left|x_{s\hat{s};t\hat{t}}^{(1\hat{1},2\hat{2}),ii}\right| + \left|x_{s\hat{s};t\hat{t}}^{(\hat{1}1,\hat{2}2),ii}\right| &\le C_x \left|\hat{s}-s\right|^2 \left|\hat{t}-t\right|^2.
    \end{aligned}
  \end{equation}
\end{remark}

\subsection{Discretization procedure}
\label{sec:discretization}
An image is a collection of pixels. For notational sake, we will consider that it can be represented as a field indexed by $\N^2$ and we denote by $\tilde{x}$ the discrete quantities. Otherwise stated we have
\begin{equation}
  \tilde{x} = \left\{x_{k;l}^i; \, k\in\{1,\dots,K\}, \, l\in\{1,\dots,L\}, \, i\in\{1,2,3\}\right\} \, ,
\end{equation}
for two constants $K,L\ge 1$. In this context the discrete first order increments are deduced from their continuous counterparts~\eqref{eq:increments-1} as
\begin{eqnarray}
 \tilde{\bx}_{k\hat{k};l\hat{l}}^{(1,2);i_1} &=& 
 \sum_{k_1=k}^{\hat{k}-1}\sum_{l_1=l}^{\hat{l}-1} \Box_{k_1,k_1+1;l_1,l_1+1} \, x^{i_1} \label{eq:x12-discrete} \\
 \tilde{\bx}_{k\hat{k};l\hat{l}}^{(\hat{1},\hat{2});i_1} &=& 
 \sum_{k_1=k}^{\hat{k}-1}\sum_{l_1=l}^{\hat{l}-1} (x_{k_1+1;l_1}^{i_1} - x_{k_1;l_1}^{i_1}) (x_{k_1;l_1+1}^{i_1} - x_{k_1;l_1}^{i_1}).  \label{eq:x-hat1-hat2-discrete}
\end{eqnarray}
As far as the equivalents of~\eqref{eq:increments-2} are concerned, the discrete form of $\bx^{(11,22)}$ is 
\begin{equation}\label{eq:double-sum}
\tilde{\bx}_{k\hat{k};l\hat{l}}^{(11,22);i_1 i_2} = \sum_{k_2=k}^{\hat{k}-1}\sum_{l_2=l}^{\hat{l}-1} 
\left(\sum_{k_1=k}^{k_2-1}\sum_{l_1=l}^{l_2-1} \Box_{k_1,k_1+1;l_1,l_1+1} \, x^{i_1} \right) \Box_{k_2,k_2+1;l_2,l_2+1} \, x^{i_2} \, .
\end{equation}
For sake of conciseness, the discrete expressions for the other increments in~\eqref{eq:increments-all} are left to the patient reader. Let us just mention that central difference approximations will be applied when considering the first-order derivatives in $d_{\hat{1}\hat{2}} x$ (see expression~\eqref{eq:conventions}).

\section{Numerical experiment}
\label{sec:numerical-experiment}
In this section, we illustrate the use of 2-d signatures as features for image texture classification. There are two ways to look at the set of features we are considering in this experiment:

\begin{enumerate}[wide, labelwidth=!, labelindent=0pt, label=(\roman*)]
\setlength\itemsep{.1in}
\item One uses a classic principal component analysis (PCA) on the image pixels, then this data is enriched with the second order increments in~\eqref{eq:increments-all}.
\item One directly uses all the increments in~\eqref{eq:increments-all}, and performs a PCA on the first order increments $\bx^{(1,2);i}$ defined by~\eqref{eq:increments-1}.
\end{enumerate}

\noindent
Since the value of each pixel can be deduced from the boundary terms $\{x_{0;t}^i, x_{s;0}^i; s \in \ots, t \in \ott\}$ and the increments $\bx^{(1,2);i}$, it is readily checked that the two approaches (\romannum{1}) and (\romannum{2}) are equivalent.

Principal component analysis is one of the classical methods in image classification problems, successfully implemented for applications such as face classification. The technique using PCA for face classification was named eigenfaces~\cite{TP91}, whose basis set for all images is  formed by principal components. In particular, the original training images may be represented by their projections on the principal components. Dimensionality reduction and feature extraction are achieved by keeping just the first few principal components, which have higher signal-to-noise ratio. Here we extend this method to texture classification, enriching our data with the natural nonlinear features given by second order signatures. We also use simple PCA features as a baseline. Let us highlight again the fact that principal components can be embedded in our signature analysis, when including both first and second order signatures in our set of features.

\subsection{Training and testing data}\label{sec:training-data}

\begin{figure}[t]
	\centering
	\includegraphics[width=0.99\linewidth]{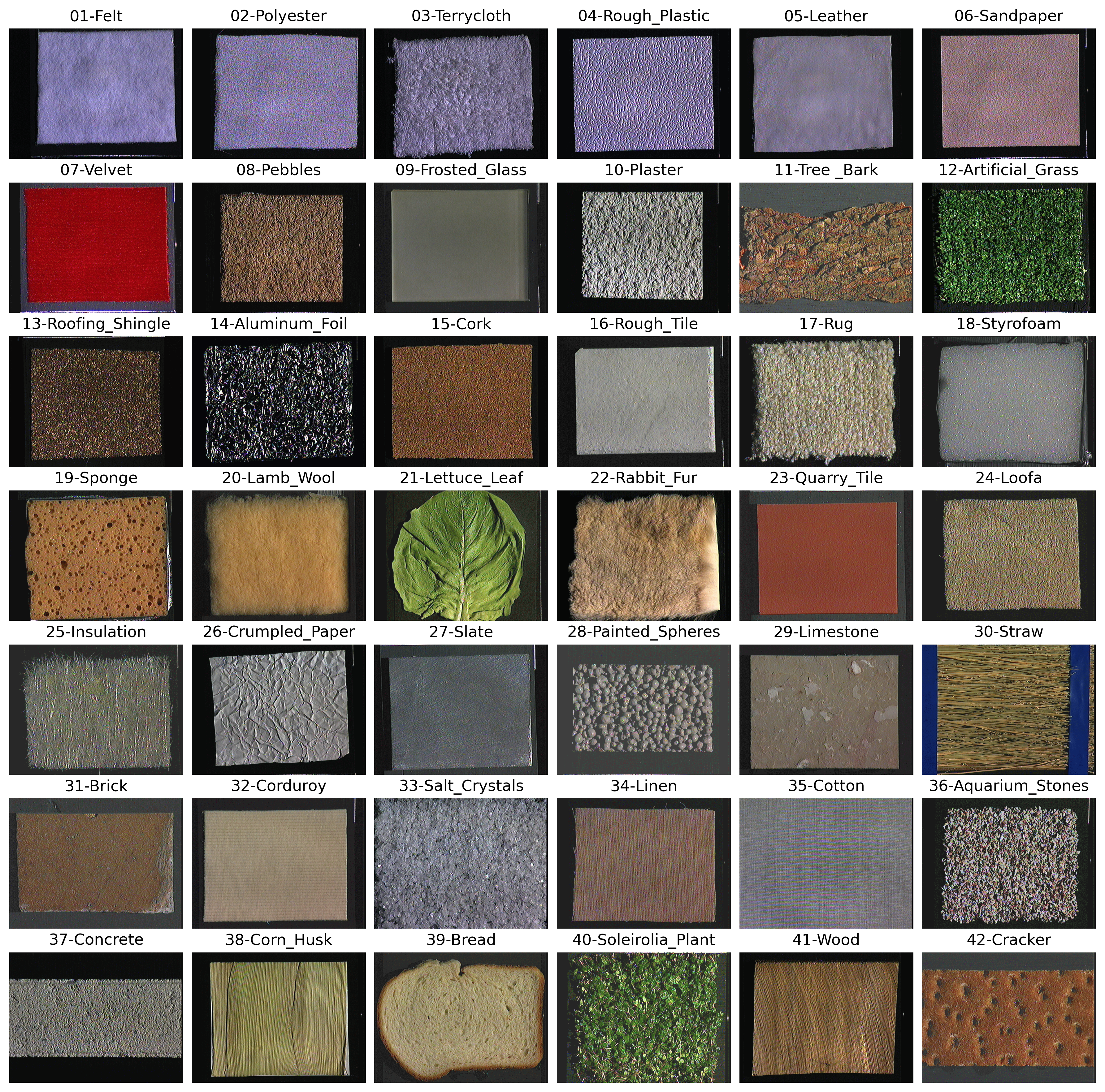}
	\caption{Forty-two different textures are used in this experiment.}
	\label{fig:all_textures}
\end{figure}

\begin{figure}[t]
	\centering
	\includegraphics[width=0.66\linewidth]{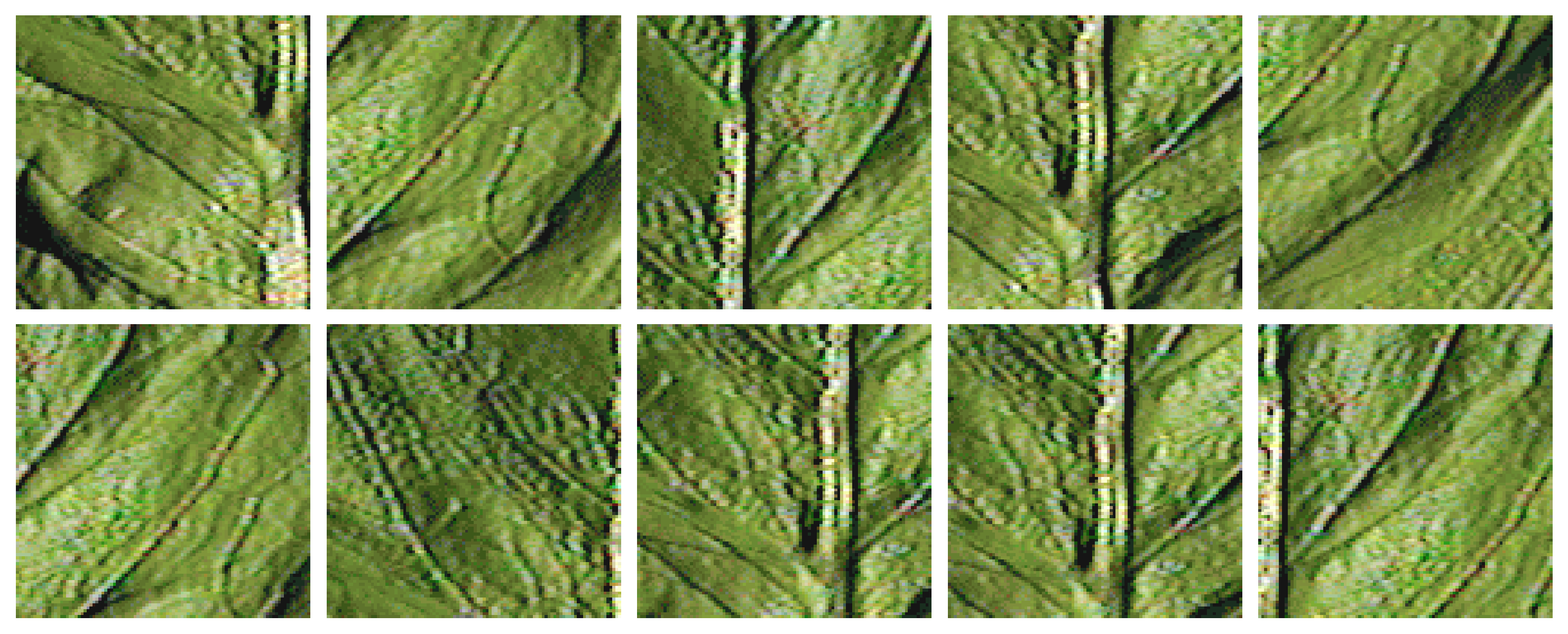}
	\caption{Ten samples from the texture ``21-Lettuce Leaf'' in Figure~\ref{fig:all_textures}.}
	\label{fig:lettuce_samples}
\end{figure}

The forty-two textures used for this experiment are depicted in Figure~\ref{fig:all_textures}. They are from a standard data set called CuRRET: Columbia-Utrecht Reflectance and Texture Database~\cite{Cur}. We randomly sample $(100 \times 100)$-sized images from each texture. Ten samples from every texture are used for training. As an illustration, the training data from the texture ``21-Lettuce Leaf'' are shown in Figure~\ref{fig:lettuce_samples}. Now, we have $10 \times 42 = 420$ different images belonging to 42 different texture categories in the training data. Since the data are easy to generate, we have used a larger size of testing data to more accurately evaluate the predictive performance of the machine learning model. Specifically, one hundred images from every texture have been sampled for testing. In total, this yields $100 \times 42 = 4200$ testing data.

\subsection{Feature engineering with symmetry}\label{sec:feature-engineering}
Feature engineering is the process of extracting useful information from raw data. If properly designed, the extracted features are of low dimension while preserving most of the important information in the data. This process can improve the performance of machine learning algorithms. As mentioned above, traditional ways of image feature engineering include PCA, which projects the data onto the first few principal components. Those components are encoded in the first order features~\eqref{eq:increments-1}.

For our experiment, we use the 2-d signatures described in Section~\ref{sec:discretization} to construct more features for image classification. For each image, we can calculate the discretized signatures using the above formulae \eqref{eq:x12-discrete}-\eqref{eq:x-hat1-hat2-discrete}-\eqref{eq:double-sum}. Since the texture should be symmetric, i.e. independent of the orientation of an image, we design the features as the average of the signatures under eight different orientations, the isometric group generated by rotation of $90^{\circ}$ and reflection. As an image has 3 channels (red, green, blue) and we are using 4 second order signatures, we have in total 12 features constructed by second order signatures for every image. We are also adding the first $N_{1}=0\text{ to }40$ principal components related to the first order increments $\tilde{\bx}_{k\hat{k};l\hat{l}}^{(1,2);i_1}$ in \eqref{eq:x12-discrete}. In the end, we are thus resorting to $N_{2}=12\text{ to }52$ features in our study. In an image processing context, this should be thought of as a low-dimensional set of features.

\subsection{Training}
We now assume that the training data of Section~\ref{sec:training-data} (including the images and their corresponding classes), as well as the features constructed in Section~\ref{sec:feature-engineering}, are given. We then train a random forest classifier using the joint features to reconstruct the corresponding classes of the training data. The classifier will then output a class label for any new input image. A good classifier should obviously be able to distinguish properly between different classes.

\subsection{Testing}
For each testing image, we calculate its features using the same formula as the training data. Then, the features are fed into the trained classifier to predict the class label. Next, we compare the predicted class labels with the true labels across all images in the testing data, and calculate the overall accuracy of the classification algorithm in identifying the true labels.

\subsection{Results}

\begin{figure}[t]
	\centering
	\includegraphics[width=0.66\linewidth]{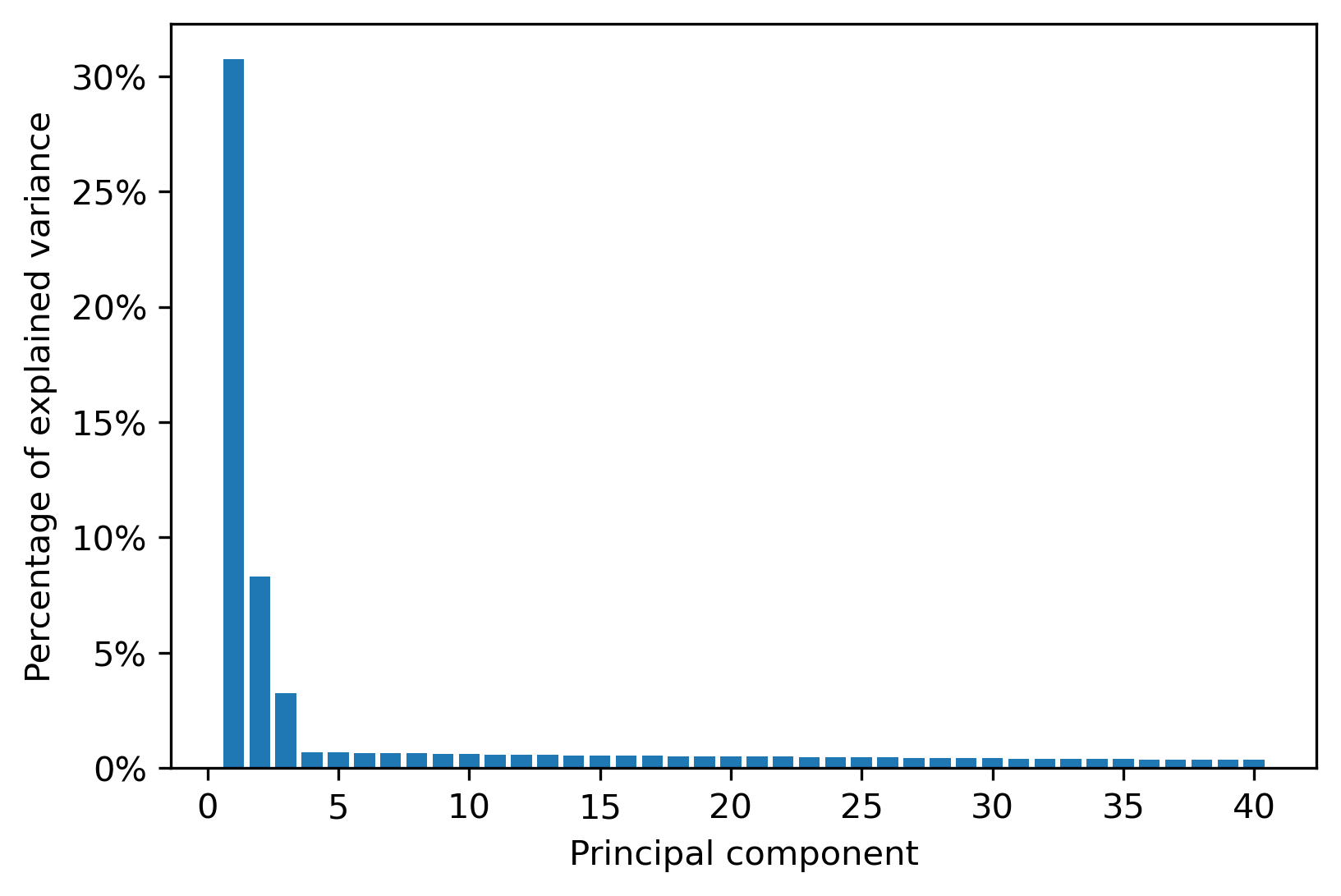}
	\caption{The percentage of explained variance of the first 40 principal components. Most of the variance is explained by the first 3 components. Principal component analysis is performed on the training data, where an image is reshaped into a vector.}
	\label{fig:PCA}
\end{figure}

\begin{figure}[t]
	\centering
	\includegraphics[width=0.99\linewidth]{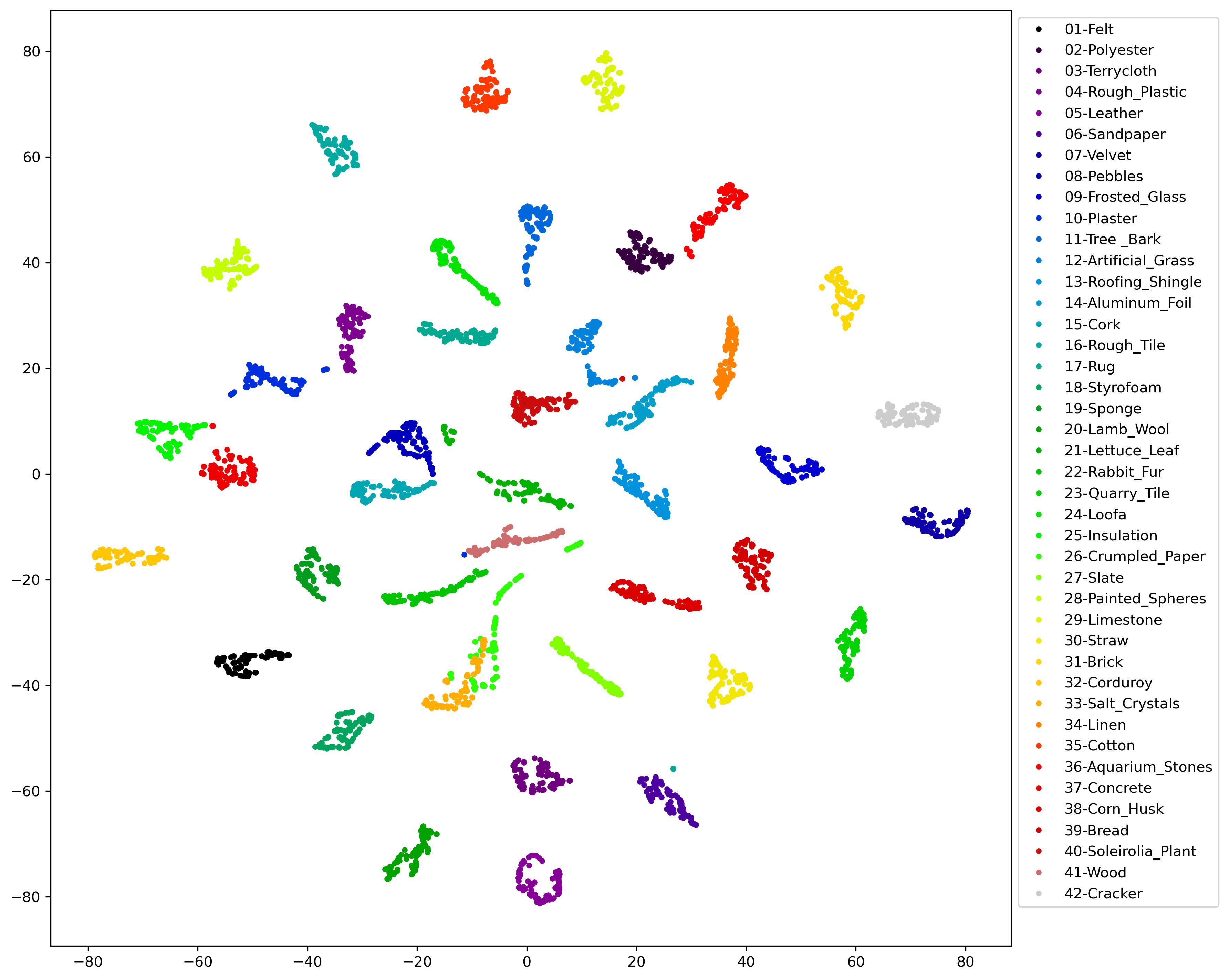}
	\caption{Two-dimensional t-distributed stochastic neighbor embedding of the 15-dimensional vectors of features of testing data. The features are generated by the first 3 principal components and 12 second order signatures. Each point represents an image. The points of the same color represent the images of the same class. The images are well differentiated by this 15-dimensional set of features, even before plugging into the classification algorithm.}
	\label{fig:TSNE}
\end{figure}

\begin{figure}[t]
	\centering
	\includegraphics[width=0.495\linewidth]{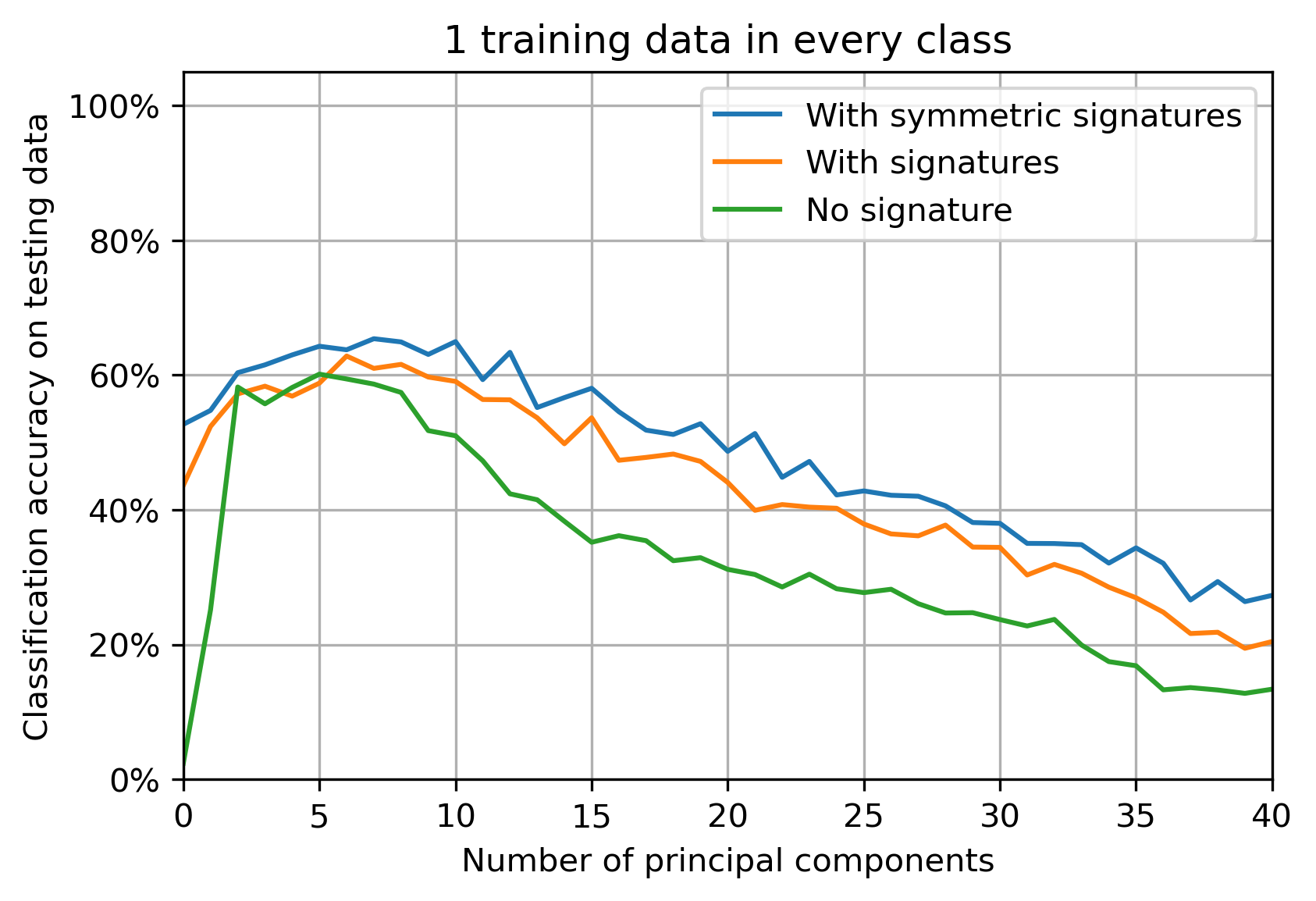}
	\includegraphics[width=0.495\linewidth]{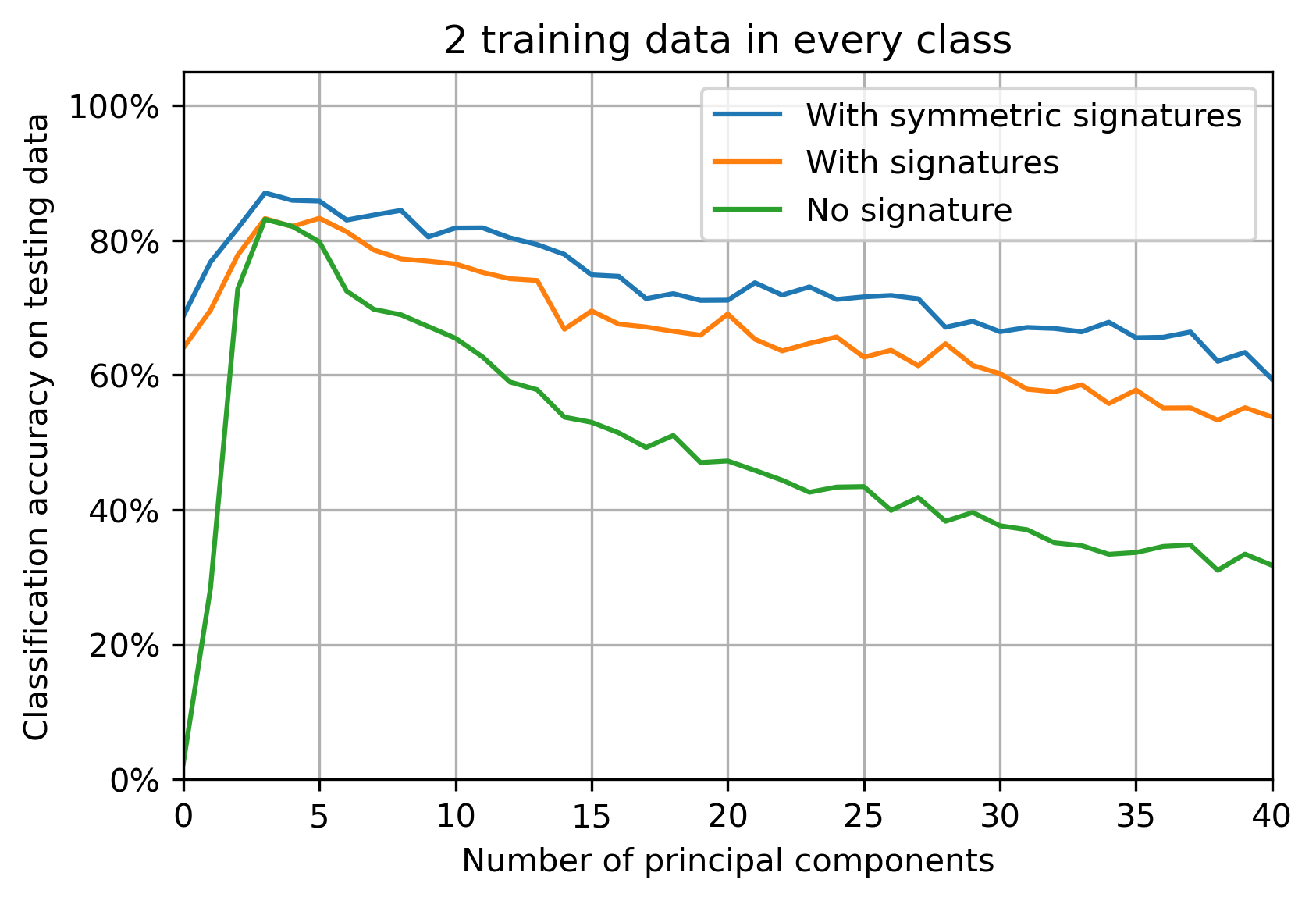} \\
	\includegraphics[width=0.495\linewidth]{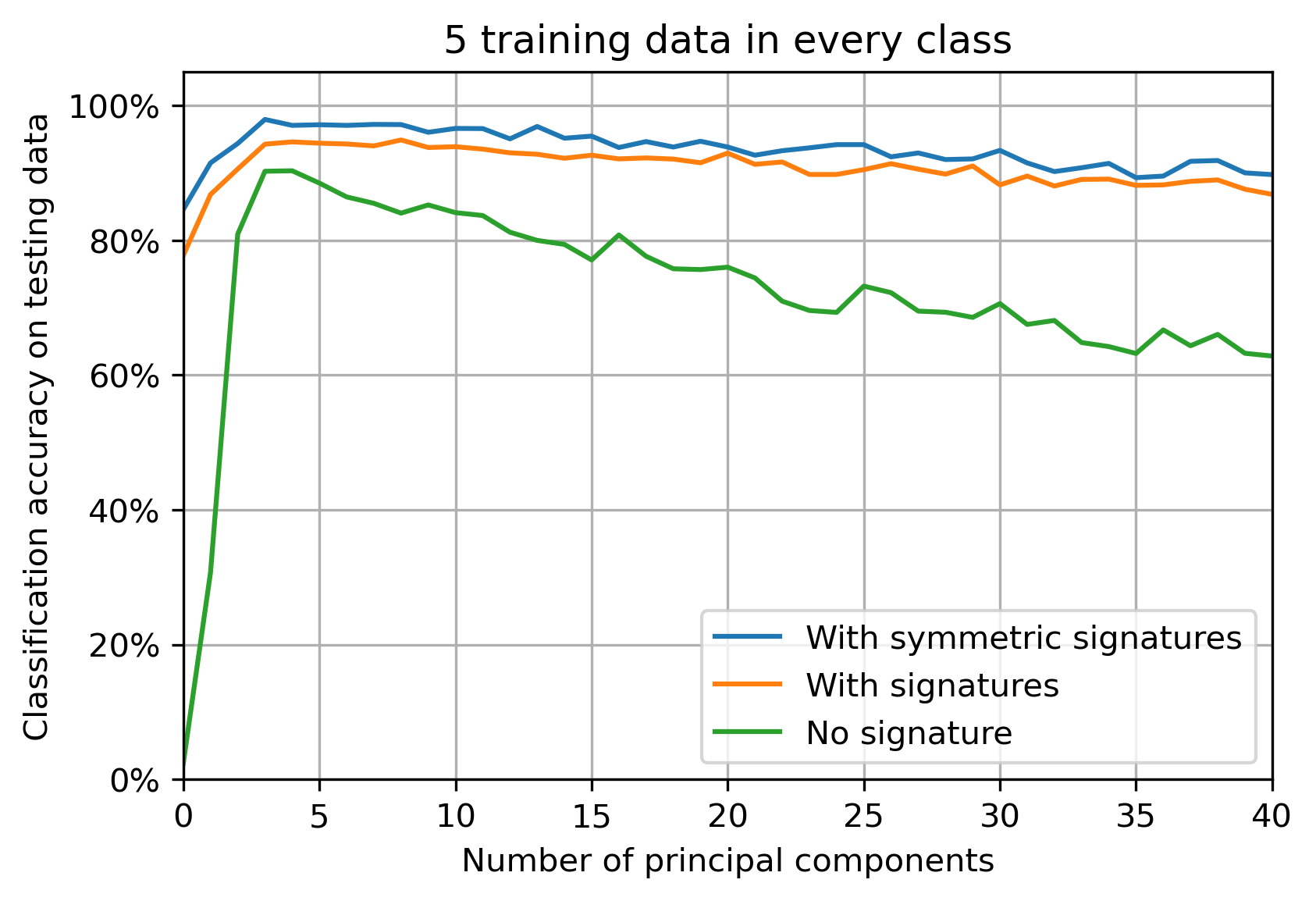}
	\includegraphics[width=0.495\linewidth]{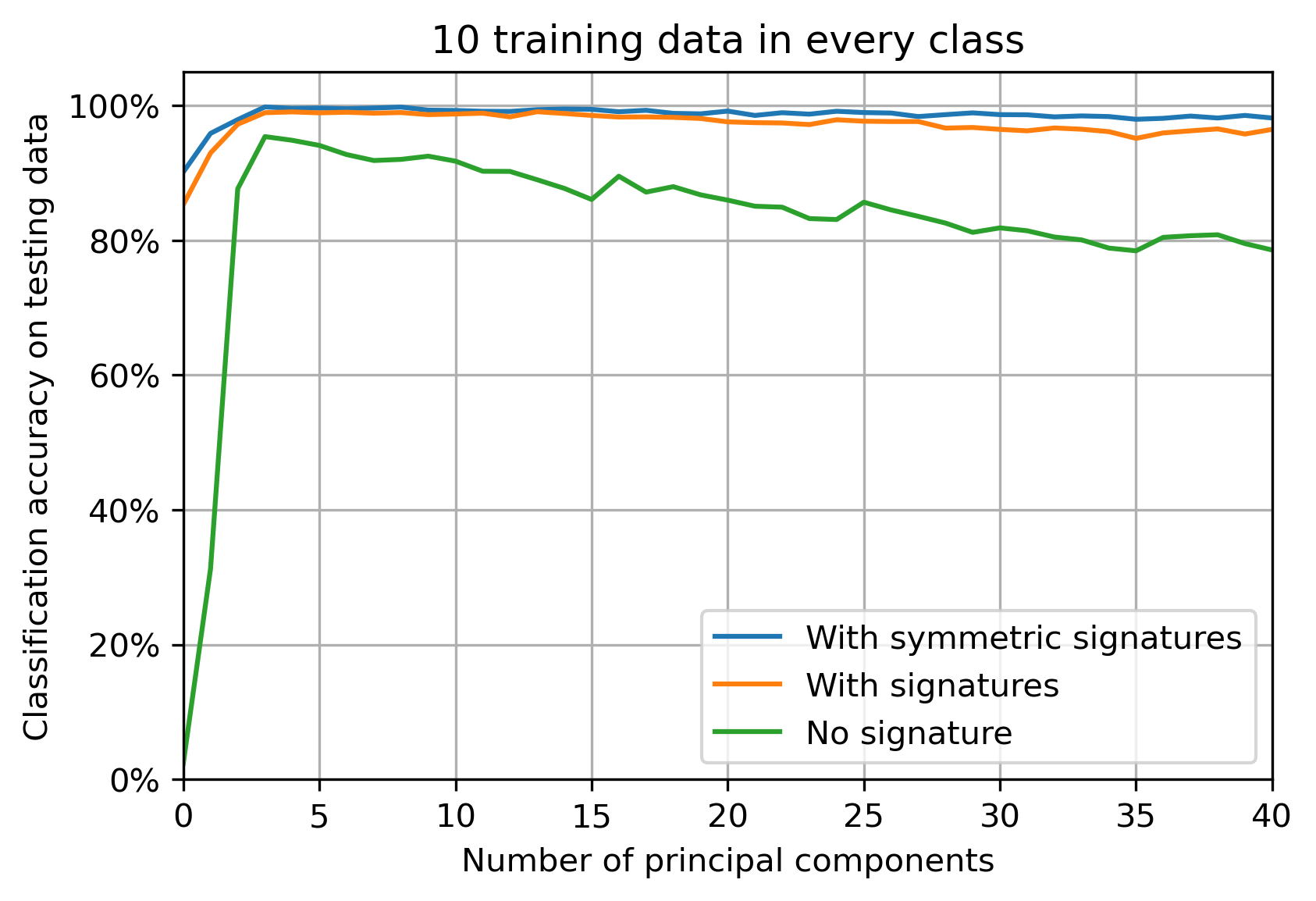}
	\caption{Classification accuracies of the random forest classifiers. When there is no signature or principal component, the accuracy is fixed at $1/42$ since there is no feature and we are predicting 1 out of 42 classes.}
	\label{fig:accuracies}
\end{figure}

Our aforementioned image classification pipeline uses features generated by principal components as well as features constructed by signatures. The percentage of explained variance of the first 40 principal components is shown in Figure~\ref{fig:PCA}. As one can see, most of the variance is explained by the first 3 components. Next to visualize the testing data in the high-dimensional feature space, we use a nonlinear dimensionality reduction technique called t-distributed stochastic neighbor embedding (t-SNE)~\cite{MH08}. T-SNE embeds high-dimensional data into two dimensional space, while trying to preserve local structures and relative distances of the data. See Figure~\ref{fig:TSNE}. The 15-dimensional vectors of features, generated by the first 3 principal components and 12 second order signatures, are embedded into a 2-dimensional space. Each point represents an image. The points of the same color represent the images of the same class. As one can see on the figure, the images are well differentiated by this 15-dimensional set of features, even before plugging into the classification algorithm.

The classification accuracies of the trained random forest classifiers on testing data are demonstrated in Figure~\ref{fig:accuracies}. We include accuracies under different feature sets. To be specific, we compare the accuracies with different numbers of principal components, as well as with and without features constructed by second order signatures.  We also contrast the classification task with or without symmetry in the construction of the signature features. Eventually, we compare the results under different sizes of training data, while the testing data keep the same. As one might expect, having more training data can improve the classification accuracy.

The numerical results suggest that when we use symmetric signatures and the first 3 principal components as features, best performance is attained. The 15-dimensional feature set yields close to 100\% classification accuracy when there are just 10 training data in every class. As nonlinear features complementary to the linear features provided by PCA, second order signatures can enhance the classification of textures. In fact, as the decay of performance when adding too many principal components suggests, not very useful features can negatively impact the machine learning classifier. On top of that, the procedure to average over 8 orientations when constructing features from signatures is helpful in improving the classification accuracy, and this is consistent under different sizes of training data.

\section{Conclusion}\label{sec:conclusion}
We have produced a low dimensional set of features, based on first and second order signatures of 2-d indexed fields. Those objects stem naturally from elementary calculus in the plane considerations. Therefore they provide a very convenient set of parameters, expressed as nonlinear functionals of a field indexed by $\R^{2}$. In addition, we have shown that 2-d signatures yield excellent performances for texture classification on a concrete example.

In view of the above observations, 2-d and more generally higher dimensional signatures certainly deserve further investigations. Below is a non exhaustive list of the research lines we wish to explore:

\begin{enumerate}[wide, labelwidth=!, labelindent=0pt, label=(\roman*)]
\setlength\itemsep{.1in}
\item
Increase the number of experiments, progressively asserting the validity of 2-d signatures as a meaningful set of characteristics for different categories of fields. Specific domains of application should include material science as well as civil engineering, where texture type features play a prominent role.

\item
Get a better grasp on the underlying algebraic structures related to 2-d signatures. This study should go beyond the investigation lead in \cite{Giu+22} (restricted to the $\dd_{\hat{1}\hat{2}} x$ differentials introduced in~\eqref{eq:conventions}), and will probably involve advanced structures such as Hopf algebras. One of the main objective in this direction is to construct all signatures (up to a given order) in a systematic way.

\item
Generalizations of rough paths notions to fields indexed by $\R^{d}$ have recently given rise to breakthroughs in the definition of singular \textsc{pde}s. Those advances have been achieved either in the landmark of regularity structures \cite{Hai14,CDOT21} or paracontrolled calculus \cite{GIP15}, and they all rely on convolution type iterated integrals. Since \textsc{pde}s are known to be related to various fundamental problems in image processing (see \cite{AK06}), it is likely that those iterated convolution integrals are also meaningful for image classification purposes. In particular the paracontrolled approach hinges on Fourier modes, which could complement the direct modes feature introduced in \eqref{eq:increments-all}.
\end{enumerate}

\noindent
As the reader can see, the signature method for image processing is a promising research direction, which deserves further investigations and improvements. Those questions will be addressed in subsequent publications.

\section*{Acknowledgments}
S. Zhang and S. Tindel are partially supported by the National Science Foundation under grant DMS-1952966. S. Zhang and G. Lin are partially supported by the National Science Foundation (DMS-1555072, DMS-1736364, DMS-2053746, and DMS-2134209), Brookhaven National Laboratory Subcontract (382247), and Department of Energy Office of Science Advanced Scientific Computing Research program (DE-SC0021142).

\bigskip
\printbibliography

\end{document}